\documentclass[10pt,twocolumn,letterpaper]{article}

\usepackage[final]{cvpr}  

\usepackage{graphicx}
\usepackage{amsmath}
\usepackage{amssymb}
\usepackage{booktabs}
\usepackage{color}
\usepackage{subfiles}
\usepackage{mmstyles}
\usepackage{standalone}
\usepackage{orcidlink}
\usepackage{float}

\usepackage{hyperref}
\hypersetup{pagebackref,breaklinks,colorlinks}

\usepackage[capitalize]{cleveref}
\crefname{section}{Sec.}{Secs.}
\Crefname{section}{Section}{Sections}
\Crefname{table}{Table}{Tables}
\crefname{table}{Tab.}{Tabs.}

\def\cvprPaperID{5252} 
\def\confName{CVPR}
\def\confYear{2025}

\begin{document}

\title{VAR-CLIP: Text-to-Image Generator with Visual Auto-Regressive Modeling}

\author{
    Qian Zhang\textsuperscript{1}~~~
    Xiangzi Dai\textsuperscript{2}~~~
    Ninghua Yang\textsuperscript{2}~~~  \\
    Xiang An\textsuperscript{2}~~~
    Ziyong Feng\textsuperscript{2}~~~
    Xingyu Ren\textsuperscript{3}~~~ \\
    \\
    \textsuperscript{1} Institute of Applied Physics and Computational Mathematics \\
    \textsuperscript{2} DeepGlint 
    \textsuperscript{3} Shanghai Jiao Tong University \\
    {\tt\small zhangqian18@iapcm.ac.cn, xiangzidai@deepglint.com}
}
\date{}

\twocolumn[{
\renewcommand\twocolumn[1][]{#1}
\maketitle
\begin{center}
\vspace{-0.15cm}
 \centering
\captionsetup{type=figure}
    \includegraphics[width=1.0\linewidth]{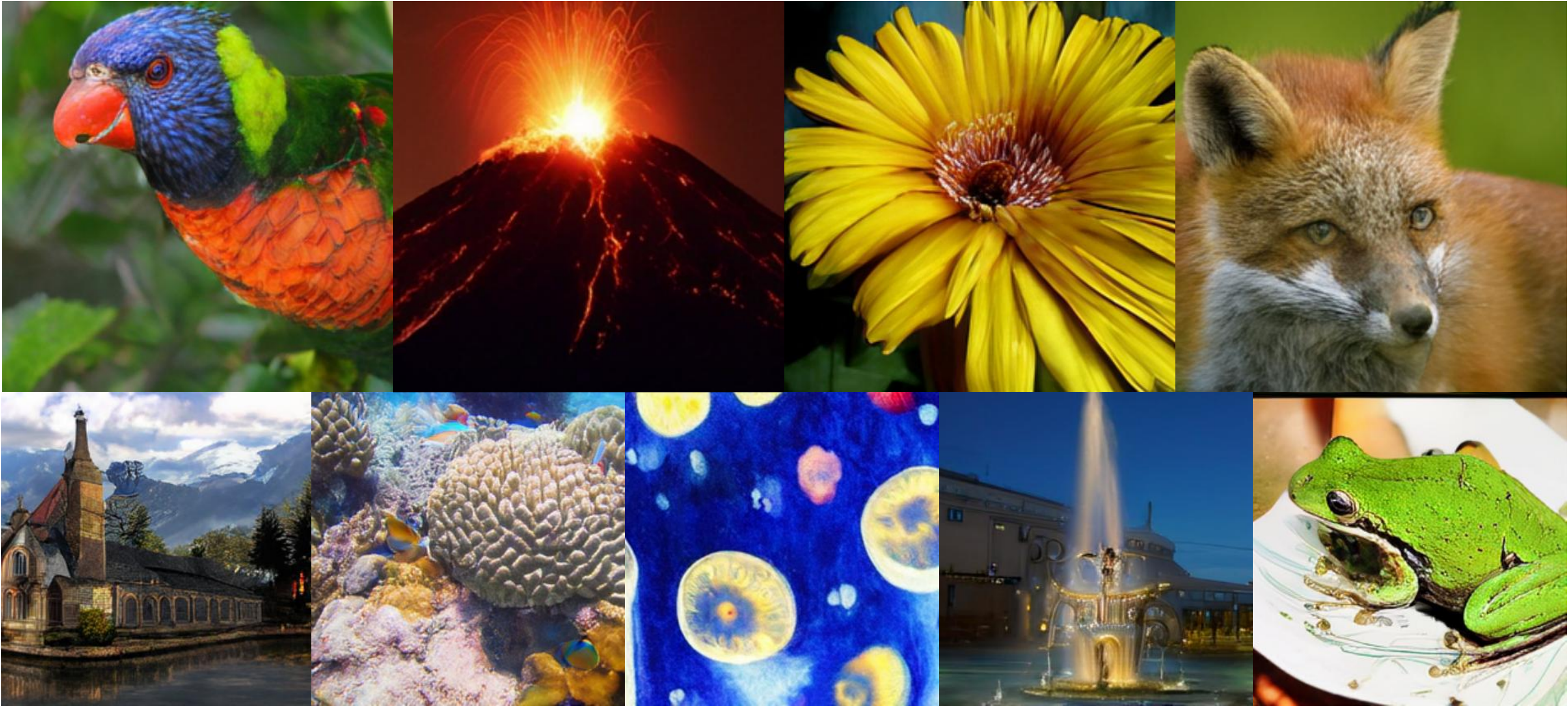}
    \caption{Exemplar images generated from text prompt by VAR-CLIP. We show $256 \times 256$ samples.\label{fig:teaser}}
\end{center}
}]

\begin{figure*}[bth]
   \centering
   \label{fig:atizzerts}
\end{figure*}

\begin{abstract}

VAR is a new generation paradigm that employs 'next-scale prediction' as opposed to 'next-token prediction'. This innovative transformation enables auto-regressive (AR) transformers to rapidly learn visual distributions and achieve robust generalization. However, the original VAR model is constrained to class-conditioned synthesis, relying solely on textual captions for guidance. In this paper, we introduce VAR-CLIP, a novel text-to-image model that integrates Visual Auto-Regressive techniques with the capabilities of CLIP. The VAR-CLIP framework encodes captions into text embeddings, which are then utilized as textual conditions for image generation. To facilitate training on extensive datasets, such as ImageNet, we have constructed a substantial image-text dataset leveraging BLIP2. Furthermore, we delve into the significance of word positioning within CLIP for the purpose of caption guidance. Extensive experiments confirm VAR-CLIP's proficiency in generating fantasy images with high fidelity, textual congruence, and aesthetic excellence. Our project page are \url{https://github.com/daixiangzi/VAR-CLIP}

\end{abstract}

\section{Introduction}
\label{sec:intro}

Text-to-image task (T2I), which aims to generate natural and realistic images while understanding textual captions, has been an engaging challenge in the computer vision community. T2I trains on large-scale data to identify data distribution and latent space within it. It then uses text embeddings as a condition to sample the latent distribution of images, achieving satisfying image generation. Text control ~\cite{zhang2023controllable, nichol2021glide, saharia2022photorealistic, zhang2023adding} and image generation ~\cite{radford2015unsupervised, kingma2013auto, ho2020denoising} 
 play crucial roles in T2I tasks.

Existing text-to-image methods primarily fall into three categories: Generative Adversarial Network (GAN)~\cite{karras2019style, karras2020analyzing}, Diffusion Model (DM)~\cite{dhariwal2021diffusion, rombach2022high}, and Auto-Regressive model (AR)~\cite{hoffmann2022training, touvron2023llama}. GANs incorporate a discriminator to regulate image generation by measuring the disparity between real and generated images~\cite{goodfellow2014generative}. While GANs excel at simplifying models, speeding up inference, and enhancing image quality, they struggle with issues like model collapse and limited diversity. DMs gradually eliminate noise from Gaussian noise to produce diverse images ~\cite{ho2020denoising, song2020denoising}. The attention module in DMs helps in aggregating essential visual concepts as outlined in the text ~\cite{rombach2022high, qu2024discriminative, zhou2024migc, wang2024instancediffusion, feng2024ranni}. DMs represent a great implementation with superior evaluation metrics due to their generation capabilities ~\cite{saharia2022palette, nichol2021glide, ho2022video, hertz2022prompt, peebles2023scalable}. However, the diffusion process incurs high computational costs and inference time owing to the iterative diffusion steps.

Auto-Regressive models (ARs) generate images by predicting the next token from a discrete prefix ~\cite{raffel2020exploring, devlin2018bert, achiam2023gpt}. The alignment between ARs and large language models (LLMs) offers a unique advantage for cross-modality fusion in textual and graphical generation. Traditional ARs like Pixel-RNN ~\cite{van2016pixel}, Pixel-CNN ~\cite{van2016conditional}, and GPT-2-like transformers ~\cite{radford2019language} have made strides in image quality and inference time. Recently, models like LlamaGen ~\cite{sun2024autoregressive}, VAR ~\cite{tian2024visual} and MAR ~\cite{li2024autoregressive} have demonstrated significant prowess in image generation, surpassing Diffusion Models (DMs). VAR introduces a novel auto-regressive model paradigm, shifting from next-token prediction to next-scale prediction, thereby enhancing computational performance and image quality. However, VAR is primarily used for class-conditional image generation, raising the question of its suitability for the T2I task.

To address this question, we introduce VAR-CLIP, a text-to-image framework that leverages CLIP to guide VAR in generating images containing textual information. VAR-CLIP adopts a two-stage training approach: first training a multi-scale VQVAE/VQGAN and then using the CLIP text encoder to extract representations of text captions as conditional tokens for image generation. We also explore how the word position in CLIP influences embeddings, noting that the first 20 tokens carry more weight than the others. Overall, our main contributions are summarized as follows:

\begin{itemize}
\item We propose a framework VAR-CLIP for high-quality text-to-image generation with minimal inference time, which use CLIP to obtain text embedding as the condition of VAR to generate images.

\item We have created a text-image pair dataset for ImageNet using BLIP-2, enabling ImageNet to support the T2I task.  

\item We investigate the importance of word position in CLIP. CLIP supports a maximum of 77 tokens, but they have the importance of imbalance for each tokens, the first 20 tokens without start of token and end of token contribute more to the caption. 
\end{itemize}

\begin{figure*}[htbp]
  \centering
   \includegraphics[width=1.0\linewidth]{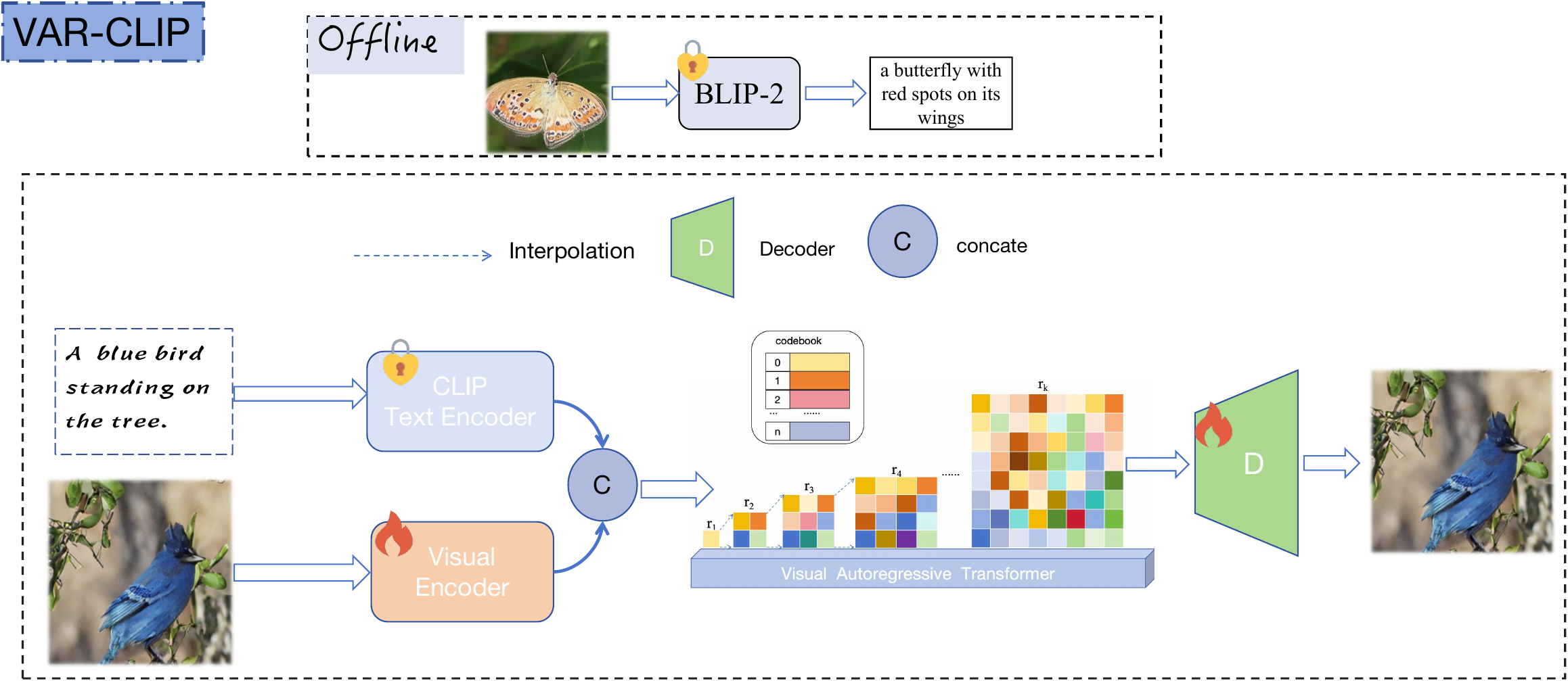}
   \caption{An illustration of VAR-CLIP. For a given text prompt and image, VAR-CLIP generates text embeddings from a pre-trained CLIP model and visual embeddings from a VAR encoder. The text embedding serves as a condition to guide the generation of multi-scale tokens and the final image. The Visual Autoregressive Transformer (VAR) generates these multi-scale tokens through next-scale prediction. During training, we utilize BLIP-2 to obtain text captions.}
   \label{fig:framework}
\end{figure*}

\section{Related Work}
\noindent\textbf{Text-to-Image Generation}

Text-to-image (T2I) generation techniques can be differentiated based on their probability distribution acquisition methods, falling primarily into three categories: Generative Adversarial Networks (GANs)~\cite{karras2020analyzing, brock2018large}, diffusion models~\cite{feng2022training}, and auto-regressive models~\cite{li2024controlvar, ma2024star}. GANs initiate the image generation process from stochastic noise, leveraging a discriminator to progressively mold the noise into coherent imagery. Despite their potential, GANs are frequently hampered by training instability. GigaGAN ~\cite{kang2023scaling} has exemplified the efficacy of GANs in T2I synthesis, delivering swift inference times, high-fidelity images, and a spectrum of latent space editing capabilities.

Diffusion models have recently gained prominence in the field of image generation, particularly for text-to-image (T2I) synthesis. Unlike GANs, these models introduce noise incrementally to create a Gaussian distribution, from which high-quality and diverse images are generated through a denoising process. Variants such as SD3.0~\cite{esser2024scaling}, SORA~\cite{videoworldsimulators2024} and DALL-E 3~\cite{betker2023improving} have demonstrated excellence in T2I tasks. However, challenges remain in improving performance, primarily due to the computational demands of multi-stage denoising and the complexity of integrating with language models.

Auto-regressive methods generate images by predicting the next token from a codebook, utilize architectures of Large Language Models (LLMs) like GPT~\cite{floridi2020gpt}, BERT~\cite{devlin2018bert}, and LLaMA~\cite{touvron2023llama, touvron2023llama}. The Vector Quantized Variational Autoencoder (VQVAE)~\cite{van2017neural} enables unsupervised, efficient, and interpretable image representation through discrete tokens. VQGAN~\cite{esser2021taming} improves upon this with transformer integration, enhancing high-resolution image quality beyond PixelCNN's capabilities. VQVAE-2~\cite{razavi2019generating} expands on this for large-scale synthesis, and Yu et al.~\cite{yu2021vector} refine efficiency and accuracy by incorporating Vision Transformers (ViTs)~\cite{dosovitskiy2020image}. Masked generation techniques~\cite{chang2022maskgit, li2023mage} have notably increased auto-regressive decoding speeds on ImageNet by 48 times.

Building on Li's observation that discrete-value spaces are unnecessary for auto-regressive models, Masked Auto-regressive Models (MARs)~\cite{li2024autoregressive} have been developed, capitalizing on the swiftness of sequence modeling. However, MARs' 1D sequence encoding may neglect 2D image spatiality. Tian's Visual Autoregressive Modeling (VAR)~\cite{tian2024visual} addresses this with a "next-scale prediction" approach, surpassing diffusion transformers in image generation quality. The integration of VAR in text-conditioned synthesis presents an opportunity for future research.

\section{Approach}
\label{sec:approach}

In this section, we introduce the details about our text-to-image generation framework and the two-stage training strategy.

In \cref{fig:framework}, our framework consists of three components: a pre-trained text encoder such as the Contrastive Language–Image model connecting text and images (CLIP), a multi-scale image tokenizer(multi-scale VQVAE) and a conditional visual autoregressive transformer (VAR). During training, a text is initially encoded as embedding $\ve_c$ by the text encoder and as the condition start tokens $\vc$, Subsequently, we generate multi-scale image tokens through a next-scale prediction strategy using VAR, and obtain residual design on $\hat{f}$ through multi-scale, Finally, we reconstruct the same image using the decoder with $\hat{f}$. Specifically, we obtain image captions using BLIP-2. 

\subsection{Pre-trained Text Encoder}
Pre-trained text encoder map the text inputs into embedding space. Contrastive Language–Image Pre-training (CLIP) ~\cite{radford2021learning}  learns visual concepts from natural language supervision, establishing a link between text and images. For any input (a text $T$, denoted as $x$), CLIP converts it into a latent embedding $\ve_c$ using the text encoder:
\begin{align}
\ve_c &= f_{\mathrm{CLIP}}(x), \mathrm{where}~x\in\{T\}. 
\end{align}

The pre-trained CLIP model is trained on 400 million images and a wide variety of natural language text. By maximize the cosine similarity of the text and image embedding, CLIP learns a multi-modal embedding space.

In text-to-image generation, limited text-image training data makes it challenging to utilize large image datasets like ImageNet. CLIP helps bridge text and image by assuming they share embeddings in a latent space. In this context, the ViT-L/14 variant of CLIP~\cite{radford2021learning} is employed for both training and inference.

\subsection{Multi-Scale Image Tokenizer}
\label{subsec:vqgan}
VQVAE~\cite{oord2017neural} transforms image into discrete image tokens to generate high-quality result. The next token can be predicted by its prefix from the transformer, contributing to the generation process. Different from VQVAE, multi-scale VQVAE employs a multi-scale quantization autoencoder to encode an image into $K$ multi-scale discrete token maps for efficient and effective generation. This multi-scale approach enhances both quality and speed. The autoencoder then assists an autoregressive transformer in predicting the next-scale prediction, further improving the generation process.

The architecture used is similar to VQVAE but differs in having a multi-scale quantization layer instead of a sequence of tokens. During encoding, an image is transformed into $K$ token maps $R = (r_1, r_2, r_3,...,r_K)$, $r_k$ ranging from small to large, $r_1$ is the start tokens with $1 \times 1$ token map, and size of $r_K$ is the $h_K \times w_K$, where $h_K \times w_K$ is the origin image size. The next-scale token depends on its prefix.

\begin{align}
    p(r_1, r_2, ..., r_k) = \prod_{k=1}^{K}{p(r_k|r_1,r_2,..., r_{k-1})} 
\end{align}

All token maps share the same codebook $\cZ \in \Rbb^{V, C}$ through optimizing encoder (E) and decoder (D) parameters. Here, $V$ represents the vocabulary, and $C$ is the vocabulary channel.

For an image $I$, start by converting it into the embedding feature map $f = E(I) \in \Rbb^{dim_z\times h\times w}$, and then, the image feature map $f$ is convert to multi-scale discrete tokens $r = (r_1,r_2,...,r_K)$, where $r \in [V]^{h \times w}$. The feature $f^(i,j)$ is mapped to the code index $r_q^{(i, j)}$ of the codebook $\cZ$ based on its nearest code in terms of Euclidean distance:
\begin{align}
    r_q^{(i, j)} = \argmin_{v\in[V]}{\lVert lookup(\cZ, v) - f^{i,j}\rVert_2}.
\end{align}

where $lookup(\cZ, v)$ means the $v$-th vector in the codebook $\cZ$.

During the training of the multi-scale autoencoder, each $\cZ$ is used to look up $r_q^{(i, j)}$ in order to acquire $z^{(i, j)}$,  for an approximation of the original image $I$, Subsequently, a reconstructed image $\hat I$ is generated by the decoder (D). The multi-scale VQVAE process can be described as follows:

\begin{align}
    z_i &= \text{lookup}(\mathcal{Z}, r_i), \\
    z_i &= \text{interpolate}(z_i, h_K, w_K), \\
    \hat{f} &= \sum_{r=1}^{K} \phi(z_r), \\
    \hat{I} &= D(\hat{f})
\end{align}

In this context, $r_i$ represents the tokens at the $i$-th scale. During reconstruction, the decoding process incorporates a residual design on $\hat f$ and utilizes it as input to the decoder for image reconstruction.

After training model, the multi-scale feature maps and codebooks serve as a multi-scale image tokenizer, with the decoder reconstructing an image from its multi-scale tokens. The multi-scale VQVAE model optimized using the loss function $\cL$:
\begin{align}
\label{eq:L_vq}
    \cL &= \lVert I-\hat{I}\rVert_2 + \lVert \hat{z} - z_q \rVert_{2} + + \lambda _p \cL_p(\hat{I})
\end{align} 
where $\cL_p(\cdot)$ represents the perceptual loss, such as LPIPS~\cite{zhang2018unreasonable}, and $\lambda _p$ denotes the weight assigned to this loss. 


\subsection{Conditional visual autoregressive transformer}
\label{subsec:autoreg}
Given a sequence of discrete tokens $x = (x_1, x_2, ..., x_n)$, where each token $x_n \in [V]$ and $V$ represents the integer vocabulary, the autoregressive transformer model predicts the highest probability of the next token as $V$. This implies that the next token $x_{n}$ is  dependent on the prefix $(x_1, x_2, ..., x_{n-1})$. While this assumption views image tokens as sentences in a left-to-right order, the reality is more intricate, and image tokens exhibit spatial relationships. Each token $x^{i,j}$ is associated with its 4 neighbors: $x^{i-1,j}$, $x^{i+1,j}$, $x^{i,j-1}$ and $x^{i,j+1}$.

The conditional autoregressive transformer is designed to predict image tokens based on its CLIP embedding. In the conditional visual autoregressive transformer (VAR)~\cite{tian2024visual}, it forecasts "next-scale tokens" instead of "next-token". By quantizing a feature map $f\in mathbb^{h \times w \times C}$ into $K$ multi-scale token maps $(r_1,r_2,...,r_K)$, each at increasingly higher resolutions of $h_k \times w_k$, culminating in $r_K$ matching the original feature map's resolution of $h \times w$. we anticipate the next scale token as:

\begin{align}
    p(\textbf{r}|\textbf{c}) &= \prod_{k}{p(r_k|r_{<k}, \ve_c)}
\end{align}

where each autoregressive scale token $r_k \in [V]^{h_k \times w_k}$ represents the token map at scale $k$, with the sequence $r_{<k}$ serving as the prefix. We incorporate the CLIP image encoder $\ve_c = f_{\mathrm{CLIP}(x)}$ as a condition to guide the next scale token. During the $k$-th autoregressive step, all distributions in $r_K$ are interdependent and will be generated in parallel.


\subsection{Training Strategy and Image Caption Generation}

\noindent This task employs the two-stage training strategy:


\textbf{First Stage} We first train a multi-scale VQVAE with the image dataset in a self-supervised manner. As mentioned in \cref{subsec:vqgan}, we utilize the pre-trained multi VQVAE model from VAR, since they have not made the training code publicly available.

\textbf{Second Stage} The conditional visual autoregressive transformer is trained at second stage. Since we have paired input-output data (embedding$\rightarrow$text), the same with autoregressive transformer(AR), our objective is to maximize the likelihood of the corresponding image token.

The maximum-likelihood of the token sequence is enforce with 
\begin{align}
\cL_{\mathrm{Transformer}} = \Ebb_{x\sim p(x)}[-\log{p(s)}]
\end{align}

\noindent \textbf{Image Caption Generation} Large datasets like ImageNet~\cite{krizhevsky2012imagenet} lack captions for each image. One approach is training with image embeddings and inference with text embeddings~\cite{wang2022clip}. However, due to the richness of image data, the generation results often lack quality and fail to meet requirements. Another method involves generating captions for training. BLIP-2~\cite{li2023blip} bridges the gap between modalities by processing both text and images. Given an image $I$, BLIP-2 effectively understands images and generates a textual description $T$. The text caption doesn't follow a template like "a photo of"; instead, it describes the image, for example, "a butterfly with red spots on its wings."

\noindent \textbf{Classifier-Free Guidance} Classifier-free guidance (cfg)~\cite{ho2022classifier} empowers generative diffusion models to produce samples of exceptionally high fidelity. Instead of relying on the gradient direction of an image classifier for sampling, this approach integrates the score estimates from both a conditional diffusion model and an unconditional model that is trained concurrently. Inspired by techniques like those used in DALL-E 2~\cite{ramesh2022hierarchical}, which occasionally sets CLIP embeddings to zero (or utilizes a learned embedding) and randomly omits the text caption during training, we have adapted a similar classifier-free guidance strategy. In our training of VAR-CLIP, we introduce randomness by occasionally replacing CLIP embeddings with Gaussian noise. This methodology has been shown to markedly improve visual quality. The inference cfg is represented as follow:

\begin{align}
e_c = (1+t)e_c - te_n
\end{align} 

where $e_c$ is the text conditional embeddings, $t$ is the weight, and $e_n$ is the Gaussian noise embedding.

\begin{figure*}[htbp]
  \centering
\includegraphics[width=1.0\linewidth]{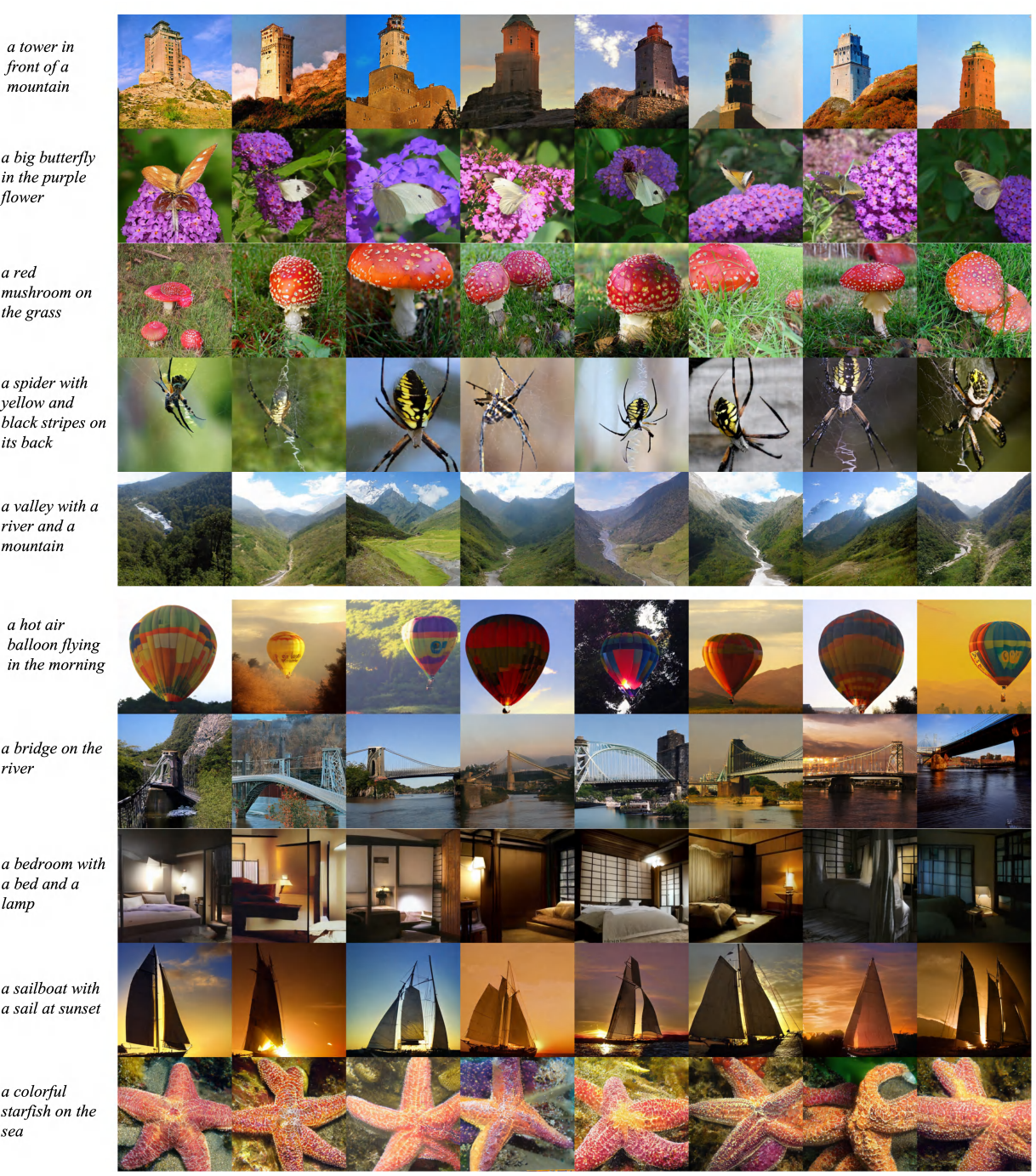}
   \vspace{0.15cm}
   \caption{Generate samples based on ten text captions trained on the ImageNet dataset, resembling those generated from BLIP-2.}
   \label{fig:style}
   \quad
\end{figure*}

\begin{figure*}[htbp]
  \centering
  \includegraphics[width=1.0\linewidth]{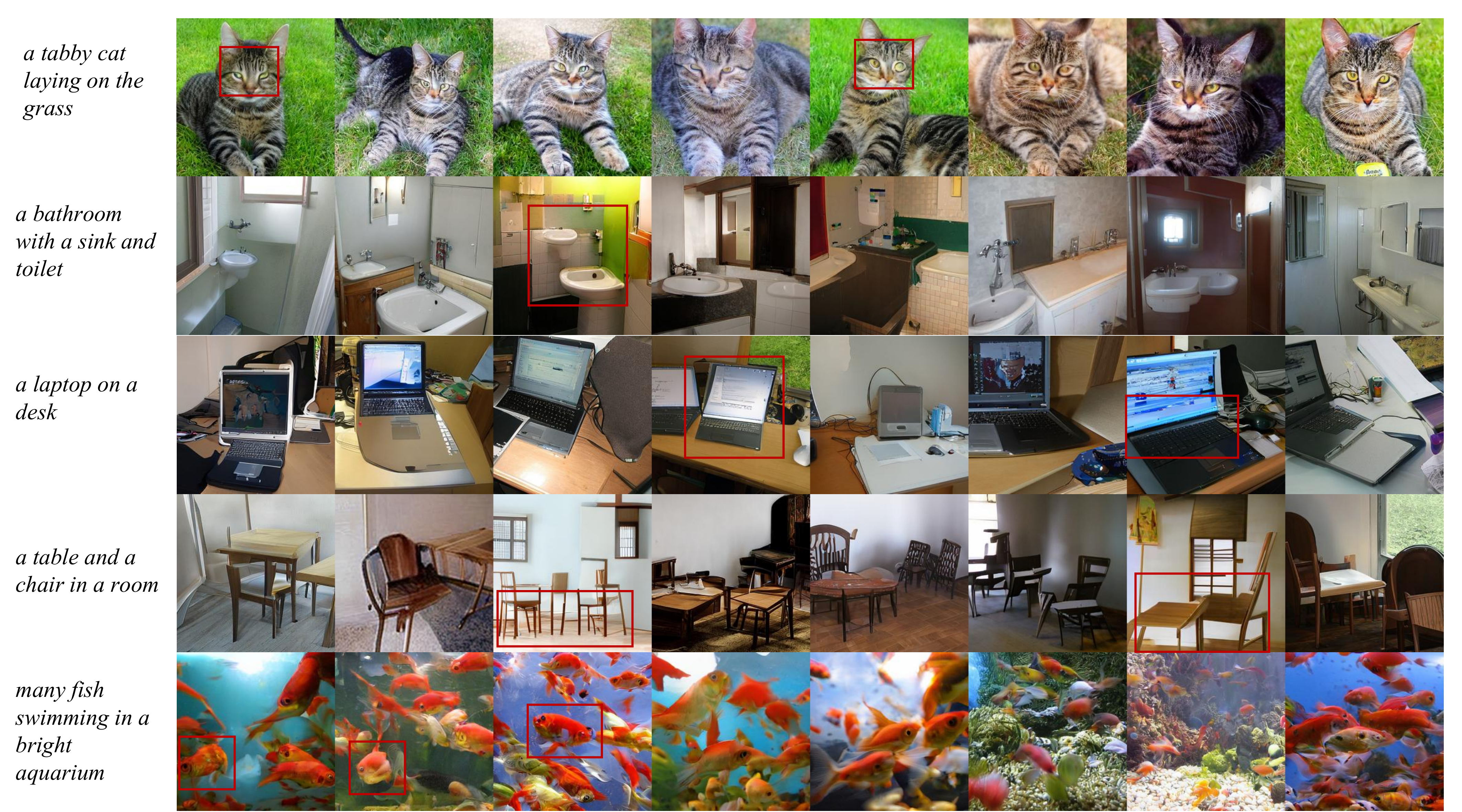}
   \vspace{0.25cm}
   \caption{Failure cases. Our method can produce noticeable artifacts in the image.}
   \label{fig:bad_style}
\end{figure*}


\begin{figure*} [htbp]
  \centering
   \begin{subfigure}[b]{0.48\textwidth}
    \includegraphics[width=\linewidth]{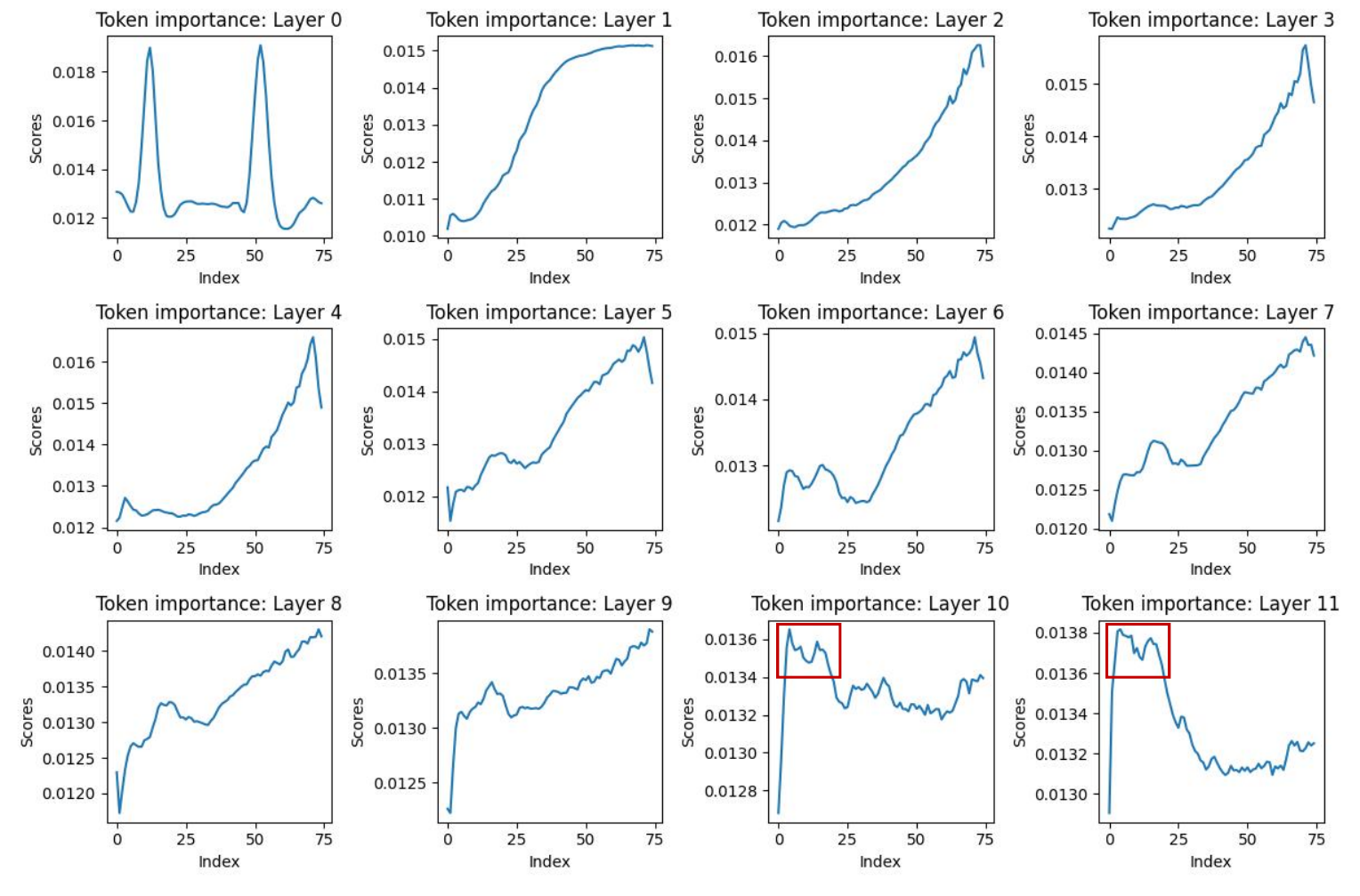}
    \caption{Clip text position score}
    \label{fig:sub1}
  \end{subfigure}
  \hfill
  \begin{subfigure}[b]{0.48\textwidth}
    \includegraphics[width=\textwidth]{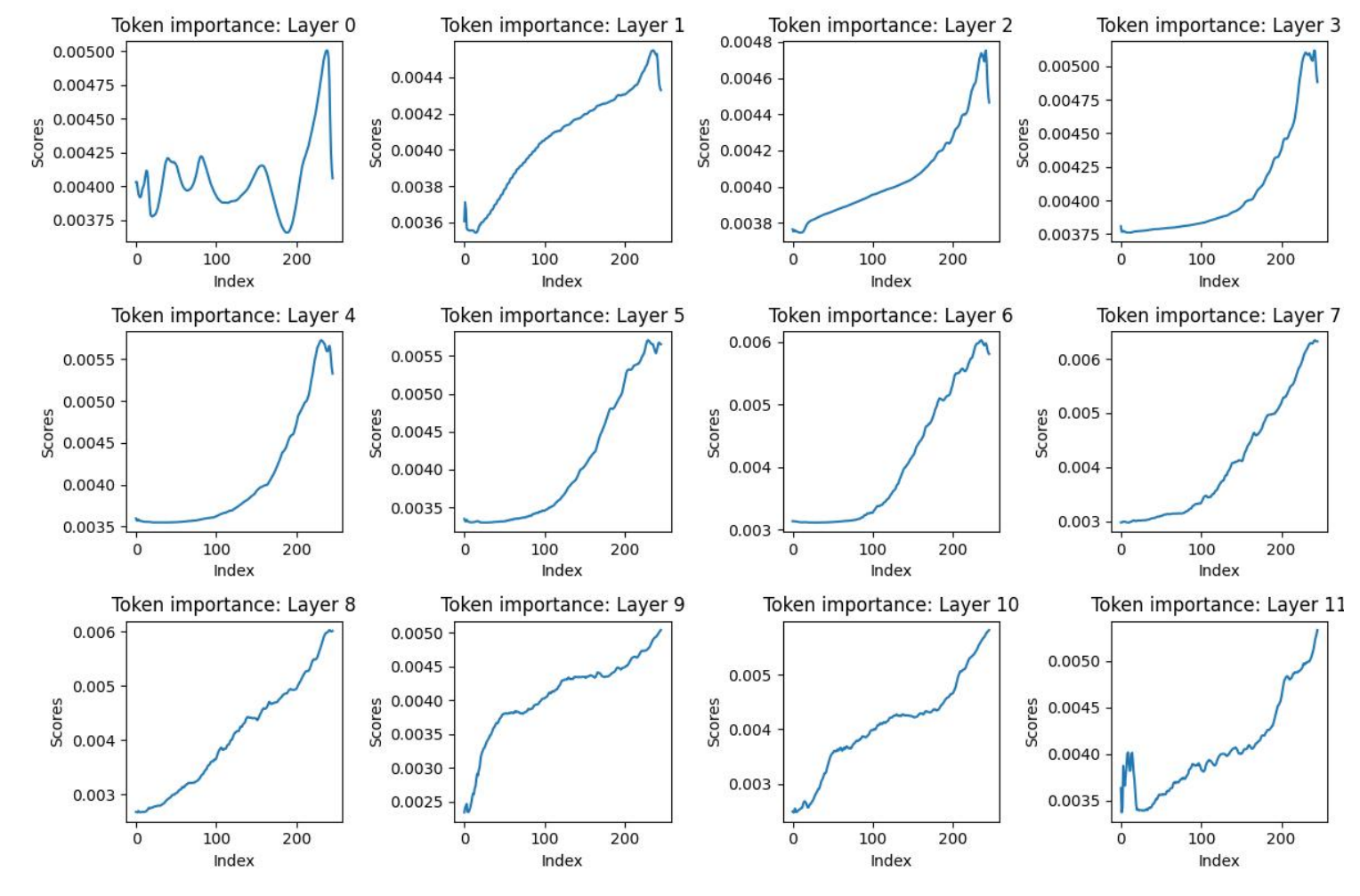}
    \caption{Long-Clip text position score}
    \label{fig:sub2}
  \end{subfigure}
  \caption{Clip position score. The different positions in a sentence have varying impacts on the weight of the sentence.}
   \label{fig:clip_position}
\end{figure*}

\section{Experiment}
\label{sec:exp}
This section describes how we evaluate our method and compare with previous approaches. First, we introduce the dataset we used and the 
implementation details of our approach.

\subsection{Dataset and Implementation Details}
\label{subsec:datasets}

\noindent \textbf{Dataset} We train and evaluate our method exclusively on the ImageNet dataset~\cite{krizhevsky2012imagenet}. ImageNet $256 \times 256$ is utilized for evaluating the conditional generation task, with 1.2 million images across 1000 classes. The entire dataset is employed to train VAR-CLIP. Captions are generated by BLIP-2, with evaluations conducted using captions styled similarly to BLIP-2, such as "a train passing through a field with a steam geyser."

\noindent \textbf{Implementation Details} We utilize a ViT-L/14~\cite{radford2021learning} variant of CLIP as a text encoder to map to an embedding space. It supports 77 tokens and has a latent channel width of 768. Our method is based on VAR, trained by Tian~\cite{tian2024visual}, with a configuration featuring $d = 16$ and 310 million parameters. Following a GPT-2~\cite{radford2019language} style transformer, we implement adaptive normalization~\cite{park2019semantic}. To incorporate classifier-free guidance~\cite{ho2022classifier}, we substitute $10\%$ of text embeddings with noise. During training, this model is trained with a learning rate of $10^{-4}$, $\beta_1 = 0.95$, $\beta_2 = 0.95$, and a decay rate of 0.05, consistent with the VAR parameters. We trained our network across 48 A100(80G) GPU machines, with a batch size of 768 per machine, over 1000 epochs, which consumed 4.151 days.

\label{subsec:quant}
\subsection{Result}

\cref{fig:style} presents qualitative outcomes, demonstrating VAR-CLIP's capability to generate images from a diverse array of textual prompts, encompassing flora, fauna, and architectural structures, as well as landscapes. VAR-CLIP adeptly renders images that align with the semantic content of the text. For instance, given the caption "a tower in front of a mountain", VAR-CLIP accurately produces an image featuring a tower with a mountain backdrop. Moreover, VAR-CLIP harnesses the text's descriptive power to evoke specific times of day. As depicted in Figure \cref{fig:style}, captions such as "a hot air balloon flying in the morning" and "a sailboat with a sail at sunset" elicit images that not only capture the essence of morning and dusk but also reflect the corresponding lighting conditions—a bright, hopeful sunrise and the soft, warm hues of a sunset. In terms of image quality, VAR-CLIP delivers high-fidelity results, offering detailed visualizations that bring the fantastical elements of the text to life. The model's proficiency in interpreting and visualizing textual nuances is evident in the rich details and clarity of the generated images.

However, our VAR-CLIP can produce failure images with noticeable artifacts in ~\cref{fig:bad_style}. This limitation is commonly observed by VAR (poor representation of animals' eyes, incomplete chairs), text-to-image tasks (fish may lose body parts when the number exceeds one), and academic datasets (e.g., ImageNet). The issue of poor representation of animals' eyes can be mitigated by using deeper networks. When $d = 30$, this problem can be controlled to generate better results. Research-driven models trained on ImageNet still exhibit significant differences in visual quality compared to commercial models trained on extensive data. Utilizing clean and information-rich datasets may lead to improved outcomes.

We explore the influence of different positions in the caption. CLIP supports 77 tokens, which include a start token and an end token. We assign the same embeddings to investigate each token's contribution to the caption. CLIP has 12 layers, and we display the token scores for each layer. In ~\cref{fig:sub1}, layer 11 indicates that the initial 25 tokens have a high score contributing to text embeddings. Long-CLIP~\cite{zhang2024long} increases the maximum input length of CLIP from 77 to 248 tokens. ~\cref{fig:sub2} shows the token scores for the 12 layers. The first 75 tokens exhibit similar trends, while the remaining tokens show an increase. Based on this, we suggest that more attention should be paid to the initial 20 tokens in the text caption.

\section{Conclusion}
In this paper, we introduce VAR-CLIP, an innovative model for text-to-image (T2I) generation. To support this framework, we have created an extensive image-text dataset using BLIP2, enhancing ImageNet's capability to facilitate T2I tasks. Furthermore, we delve into the significance of word positioning in CLIP for image generation, demonstrating VAR-CLIP's ability to produce fantasy images characterized by high fidelity, textual congruence, and aesthetic excellence.

Despite these advancements, VAR-CLIP faces certain limitations, such as the precision of captioning and the alignment of the text encoder model with the image generation process. To enhance captioning, we propose the adoption of a more sophisticated image description model capable of producing more detailed captions. Addressing text-image alignment in auto-regressive models requires a dual-pronged approach: utilizing advanced language models for text comprehension and conducting focused research akin to that in diffusion models to improve alignment.

Looking ahead, we aim to tackle the limitations identified. Our future work will focus on generating high-quality captions and aligning complex textual and visual elements, including color, spatial arrangement, and associated objects, to further refine the T2I synthesis process.

{\small
\bibliographystyle{ieee_fullname}
\bibliography{main}
}

\newpage

\end{document}